\documentclass[11pt,a4paper]{article}
\usepackage[acceptedWithA]{tacl2021v1}
\usepackage{times,latexsym}
\usepackage{url}
\usepackage[T1]{fontenc}

\usepackage[utf8]{inputenc}
\usepackage{microtype}
\usepackage{inconsolata}

\usepackage{graphicx,xcolor}  
\usepackage{pxrubrica}        
\usepackage{url}
\usepackage{booktabs}
\usepackage{adjustbox}
\usepackage{bm}
\usepackage{CJKutf8}
\usepackage{multirow}
\usepackage{amsmath,amsfonts,amssymb,mathtools}

\usepackage{tacl2021v1}

\usepackage{xspace,mfirstuc,tabulary}

\newif\iftaclinstructions
\taclinstructionsfalse 
\iftaclinstructions
\renewcommand{\confidential}{}
\renewcommand{\anonsubtext}{(No author info supplied here, for consistency with
TACL-submission anonymization requirements)}
\newcommand{\instr}
\fi

\iftaclpubformat 

\else

\fi

\newcommand{\ja}[1]{\begin{CJK}{UTF8}{ipxm}#1\end{CJK}}

\newcommand{\ulblue}[1]{\textcolor{blue}{\underline{\color{black}{#1}}}}
\DeclareMathOperator*{\argmax}{arg\,max}

\title{Graph-Structured Trajectory Extraction from Travelogues}

\author{
\textbf{Aitaro Yamamoto}$^{\clubsuit,}$\Thanks{Currently working at a company.} \hspace{0.2cm} 
\textbf{Hiroyuki Otomo}$^{\heartsuit}$ \hspace{0.2cm} 
\textbf{Hiroki Ouchi}$^{\clubsuit,\heartsuit,\blacklozenge,}$\Thanks{Corresponding author.} \\
\textbf{Shohei Higashiyama}$^{\spadesuit,\clubsuit}$  \hspace{0.2cm}
\textbf{Hiroki Teranishi}$^{\blacklozenge,\clubsuit}$ \hspace{0.2cm} 
\textbf{Hiroyuki Shindo}$^{\diamondsuit}$ \hspace{0.2cm} 
\textbf{Taro Watanabe}$^{\clubsuit}$\\
  {$^\clubsuit$ NAIST} \hspace{0.3cm} 
  {$^\spadesuit$ NICT} \hspace{0.3cm} 
  {$^\blacklozenge$ RIKEN} \hspace{0.3cm} 
  {$^\heartsuit$ CyberAgent, Inc.} \hspace{0.3cm} 
  {$^\diamondsuit$ MatBrain, Inc.} \\
  \texttt{yamamoto.aitaro.xv6@is.naist.jp}, \hspace{0.25cm}
  \texttt{otomo\_hiroyuki@cyberagent.co.jp}, \hspace{0.25cm} \\
  \texttt{hiroki.ouchi@is.naist.jp}, \hspace{0.25cm}
  \texttt{shohei.higashiyama@nict.go.jp}, \\
  \texttt{hiroki.teranishi@riken.jp}, \hspace{0.25cm} 
  \texttt{hshindo@matbrain.jp}, \hspace{0.25cm} 
  \texttt{taro@is.naist.jp}}

\date{}

\begin{document}
\maketitle
\begin{abstract}
Previous studies on \textit{sequence}-based extraction of human movement trajectories have an issue of inadequate trajectory representation. Specifically, a pair of locations may not be lined up in a sequence especially when one location includes the other geographically.
In this study, we propose a \textit{graph} representation that retains information on the geographic hierarchy as well as the temporal order of visited locations, and have constructed a benchmark dataset for graph-structured trajectory extraction.
The experiments with our baselines have demonstrated that it is possible to accurately predict visited locations and the order among them, but it remains a challenge to predict the hierarchical relations.
\end{abstract}

\section{Introduction}
Travelogues are a significant source for analyzing human traveling behavior in various fields, including tourism informatics~\cite{akehurst2009user,hao2010} and human geography~\cite{cooper2011mapping,caselli_sprugnoli_dnt2021}, because of their rich geographic and thematic content, which gives people, for example, a simulated experience of trip~\cite{haris2021extraction}.
In particular, human traveling trajectories play a central role in characterizing each travelogue, and thus, automatic trajectory extraction from travelogues is highly desired for tourism services, such as travel planning and recommendation~\cite{pang2011}.

\begin{figure}[t]
\centering
  \includegraphics[width=7.6cm]{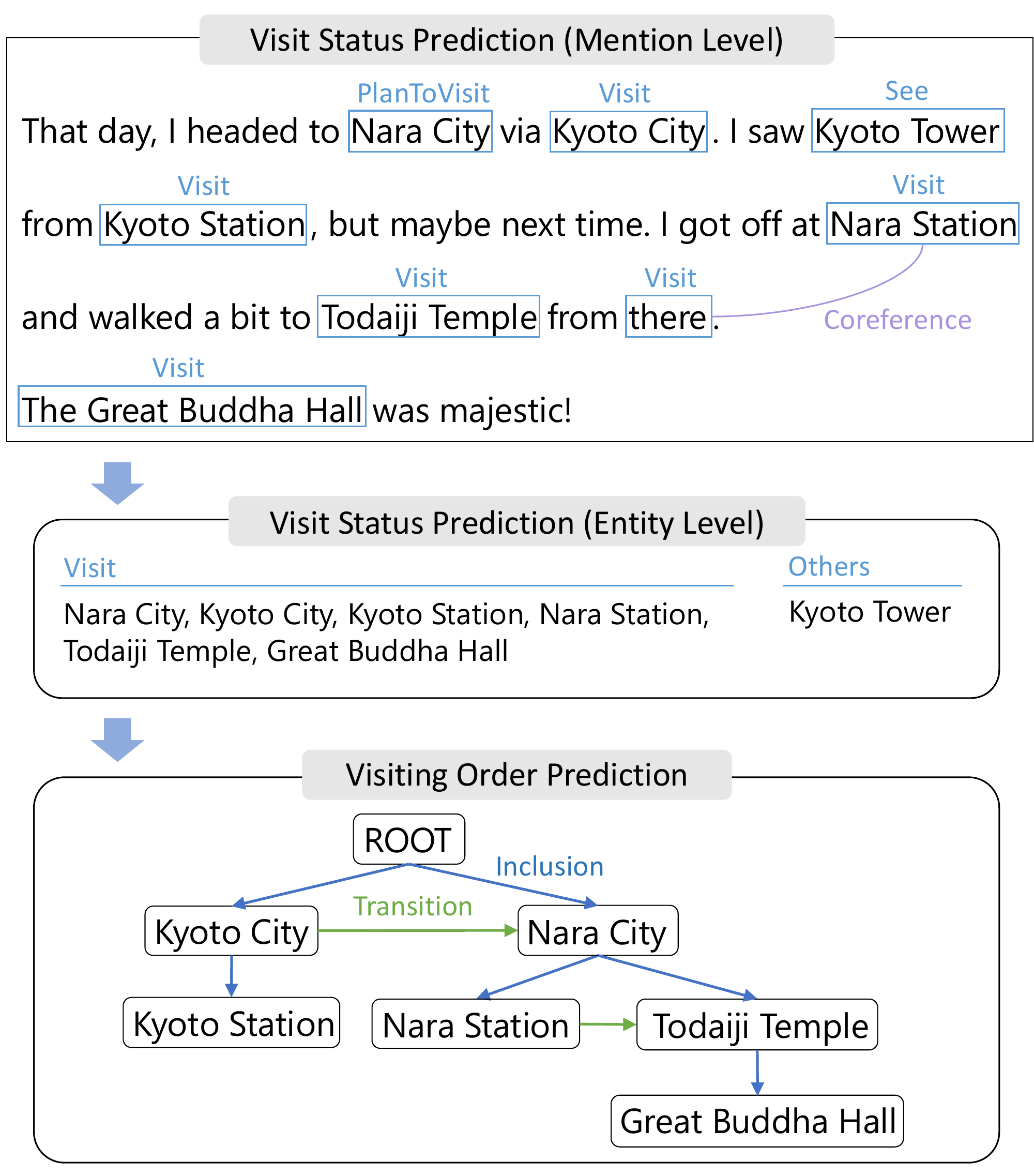}
  \caption{Illustration of our proposed tasks: visit status prediction (VSP) and visiting order prediction (VOP). The goal of VSP is to assign visit status labels to mentions (top) and to entities (middle). The goal of VOP is to build a visiting order graph by identifying inclusion and transition relations between entity pairs (bottom).}
  \label{fig:step4_annotation_overview}
\end{figure}

Some studies have addressed automatic trajectory extraction from text~\cite{ishino-2012-extracting,wagner-etal-2023-event,kori-2006-blog}.
However, these studies have two problematic issues: (i) inadequate trajectory representation and (ii) the scarcity of benchmark datasets.

First, the previous studies treated each trajectory as a \textit{sequence} of visited locations~\cite{ishino-2012-extracting,wagner-etal-2023-event,kori-2006-blog}, but a sequence is inadequate as a representation of trajectories.
This is because a pair of locations may not be lined up in a single sequence especially when one location includes the other geographically, for example, ``{Kyoto City}'' and ``{Kyoto Station}.''

Second, the previous studies constructed and used their in-house datasets for evaluating their systems, and no public text datasets have been released with trajectory information annotation.
However, shared benchmark datasets are necessary for facilitating fair comparisons with other studies and accelerating the accumulation of research findings~\cite{osuga2021}.


For the first issue, we propose a \textit{visiting order graph} illustrated at the bottom of Figure~\ref{fig:step4_annotation_overview}.
This graph has location or geo-entity nodes and relation edges between nodes, and can represent not only temporal \textit{transition relations} but also geographic \textit{inclusion relations} between visited locations.
For facilitating automatic graph construction for each travelogue, we introduce subtasks of trajectory extraction: Visit Status Prediction (VSP) and Visiting Order Prediction (VOP), as shown in Figure~\ref{fig:step4_annotation_overview}.
VSP assigns \textit{visit status labels} to mentions and entities, while VOP identifies inclusion and transition relations between nodes of the ``visited'' entities. 

For the second issue, we have constructed a dataset for training and evaluating trajectory extraction systems: {A}rukikata {T}ravelogue {D}ataset with {V}isit {S}tatus and Visiting {O}rder Annotation (ATD-VSO).\footnote{We will release our dataset at \texttt{ANNONYMIZED\_URL}.}
Our dataset comprises 100 travelogue documents annotated with the corresponding visiting order graphs, including 3,354 geo-entities (nodes) and 3,369 relations (edges) in total.

Using this dataset, we have trained and evaluated baseline systems.
Notable findings through the experiments are (i) that the systems can achieve relatively high accuracy for predicting visit status labels and the transition relations, and (ii) that the systems failed to accurately predict inclusion relations.
The latter implies an important future investigation, i.e., how to inject the knowledge of geographic hierarchical structure into the systems.

\paragraph{Contributions}
For the purpose of building a foundation for future studies, we have made two main contributions: (i) the proposal of a visiting order graph and (ii) the construction of a benchmark dataset for the trajectory extraction.\footnote{Our contributions are in the data resource direction, not the technical one such as algorithm and model sophistication. On top of the resource, we will make technical contributions in the future.}
We will release our code and dataset for research purposes.
We expect that our dataset will foster continued growth in the trajectory extraction research.

\section{Preliminaries for Data Construction} \label{sec:premise}
Our dataset, ATD-VSO, has been constructed on the basis of ATD-MCL~\cite{higashiyama-etal-2023-atd},\footnote{\url{http://github.com/naist-nlp/atd-mcl}} which is a Japanese travelogue dataset annotated with geo-entity information, including mentions and their coreference relations, using a subset of the original travelogues, the Arukikata Travelogue Dataset~\cite{arukikata-2022,ouchi-etal-2023-atd}.\footnote{\url{https://www.nii.ac.jp/dsc/idr/arukikata/}}
Annotated mentions in ATD-MCL include proper nouns (e.g., ``Nara station''), general noun phrases (e.g., ``the station''), and deictic expressions (e.g., ``there'') that refer to various types of locations, such as geographic regions, facilities, and landmarks. 
Moreover, a set of mentions that refer to the same location constitutes a coreference cluster or geo-entity.\footnote{In this paper, we use the term ``geo-entity'' or ``entity,'' instead of ``coreference cluster.''}
Given such annotated travelogues, we focus on annotating the visit status and visiting order of candidate geo-entities.

\section{Visit Status Prediction} \label{sec:vsp_task}
In this section, we describe the task definition, annotation data construction, and our baseline system for Visit Status Prediction (VSP), which is the first step of trajectory extraction with given mentions and entities.
The goal of VSP is to identify locations that the traveler\footnote{We assume that the traveler in a travelogue usually corresponds to the writer of the travelogue.} visited by assigning a visit status for each location mentioned in the travelogue.
For example, it is possible to judge that the traveler actually visited the station from the description of the real experience: ``Arrived at Kintetsu Nara Station!''
In contrast, the following factual statement does not indicate whether the traveler visited these stations or not: ``JR Nara Station is a little far from Kintetsu Nara Station.''

\subsection{Annotation Data Construction} \label{sec:vsp_ann}


We defined two types of visit status labels for entities and six types of visit status labels for mentions in Table~\ref{tab:visit_labels_for_mentions}.\footnote{Actual examples are listed in Table~\ref{tab:visit_labels_example} in Appendix~\ref{sec:app_label_example}.}
The mention labels serve to distinguish detailed status for the mentioned locations based on the context, which typically corresponds to the sentence where the mention occurs.
The entity labels serve to determine whether the traveler eventually visited the location, considering the context of the entire document.

As annotation work, native Japanese annotators at a data annotation company assigned visit status labels to each mention and entity in ATD-MCL travelogues using the brat annotation tool~\cite{stenetorp-etal-2012-brat},\footnote{\url{https://github.com/nlplab/brat}} according to the label definitions and annotation guideline.



\begin{table}[t]
\centering
\small
\begin{tabular}{llp{4cm}}
\toprule
& Label & Definition \\ \midrule
1 & \texttt{Visit} & A visit to the location is stated or implied.\\
2 & \texttt{Other} & Not 1. \\
\midrule
1 & \texttt{Visit} & The same as the label \texttt{Visit} for entities. \\
2 & \texttt{PlanToVisit} & It mentions a plan to visit the location during 
 this trip (described in the travelogue). \\
3 & \texttt{See} & Not any of 1--2, and that the traveler saw the location can be identified. \\
4 & \texttt{Visit-Past} & Not any of 1--3, and it mentions having visited the location before this trip. \\
5 & \texttt{Visit-Future} & Not any of 1--3, and it mentions the intention to visit the location after this trip. \\
6 & \texttt{UnkOrNotVisit} & The visit to the locations cannot be identified from the descriptions, or the non-visit can be identified. \\
\bottomrule
\end{tabular}
\caption{Visit status labels for entities (top) and mentions (bottom).}
\label{tab:visit_labels_for_mentions}
\end{table}

\paragraph{Inter-Annotator Agreement}
We requested two annotators to independently annotate five documents.
We then measured the inter-annotator agreement (IAA) using F1-score and Cohen's Kappa $\kappa$.
The obtained scores suggest the high agreement: F1 score of 0.80 and $\kappa$ of 0.68 for 180 mentions, and F1-score of 0.89 and $\kappa$ of 0.81 for 124 entities.

\begin{table}[t]
\centering
\small
\begin{tabular}{lrrrrrr}
\toprule
Set & \#Doc & \#Sent & \#Men & \#Ent & \#Inc+Tra\\
\midrule
Train\!\! & 70 & 4,254 & 3,782 & 2,339 & 2,343\\
Dev   & 10 &   601 &   505 &   316 &   329\\
Test  & 20 & 1,469 & 1,102 &   699 &   697\\
\midrule
Total\!\! & 100 & 6,324 & 5,389 & 3,354 & 3,369\\
\bottomrule
\end{tabular}
\caption{Basic statistics of the ATD-VSO: the numbers of documents (``\#Doc''), sentences (``\#Sent''), mentions (``\#Men''), entities (``\#Ent''), and inclusion and transition relations (``\#Inc+Tra'').}
\label{tab:stats}
\end{table}

\paragraph{Data Statistics}
95 documents were allocated among multiple annotators, and one annotator annotated each document.
The total became 100 documents, including the aforementioned five documents, with total 3,354 entities.
For the constructed dataset, the basic statistics and detailed statistics are shown in Table~\ref{tab:stats} and Table~\ref{tab:men_stats} in Appendix~\ref{sec:app_stat}, respectively.

\subsection{Task Definition}
Entity-level and mention-level VSP are defined as follows:
Given a set of entities $\mathcal{E}$ in an input document, entity-level VSP requires a system to assign an appropriate visit status label $y \in \mathcal{L}_e$ for each entity $e_q \in \mathcal{E}$.
Similarly, given an entity (or coreference cluster) $e_q = \{m^{(q)}_1, \dots, m^{(q)}_{|e_q|} \}$, which consists of one or more mentions, mention-level VSP requires a system to assign an appropriate visit status label $y \in \mathcal{L}_m$ for each mention  $m^{(q)}_i \in e_q$.

\subsection{Baseline System} \label{sec:vsp_baseline}
As our baseline system, we employ a two-step method that first predicts mention labels and then predicts entity labels based on the mention labels.
Specifically, we calculate the label probability distribution $P(y | m^{(q)}_i)$ for each mention $m^{(q)}_i \in e_q$, and select the most probable label $\hat{y}^{(q)}_i$:
\begin{eqnarray*}\label{eq:vsp_inference}
    \hat{y}^{(q)}_i = \argmax_{y \in \mathcal{L}_m} \: P(y | m^{(q)}_i).
\end{eqnarray*}

\noindent
Then, we select a label for each entity $e_q$ according to the following mention label aggregation (MLA) rules.
\begin{enumerate}
    \setlength{\parskip}{0cm} 
    \setlength{\itemsep}{0cm}
    \item If \texttt{Visit} or \texttt{PlanToVisit} has been assigned to at least one mention in $e_q$, then \texttt{Visit} is assigned to $e_q$.
    \item Otherwise, \texttt{Other} is assigned to $e_q$.
\end{enumerate}

\paragraph{Model Implementation}
As the implementation of a model for mention label prediction, we used \texttt{LukeForEntityClassification} in Hugging Face Transformers\footnote{\url{https://huggingface.co/docs/transformers/index}} with the inputs of the sentence containing the mention of interest and the position (character offsets) of the mention.

\section{Visiting Order Prediction} \label{sec:vop_task}
In this section, we first define a visiting order graph, which is a key concept for our structured trajectory extraction.
Then, we describe the task definition, annotation data construction, and our baseline system for Visiting Order Prediction (VOP), which involves prediction of geographic and temporal relations between visited locations.

\subsection{Visiting Order Graph} \label{sec:visit_graph}

\begin{figure}[t]
\centering
    \includegraphics[width=7cm]{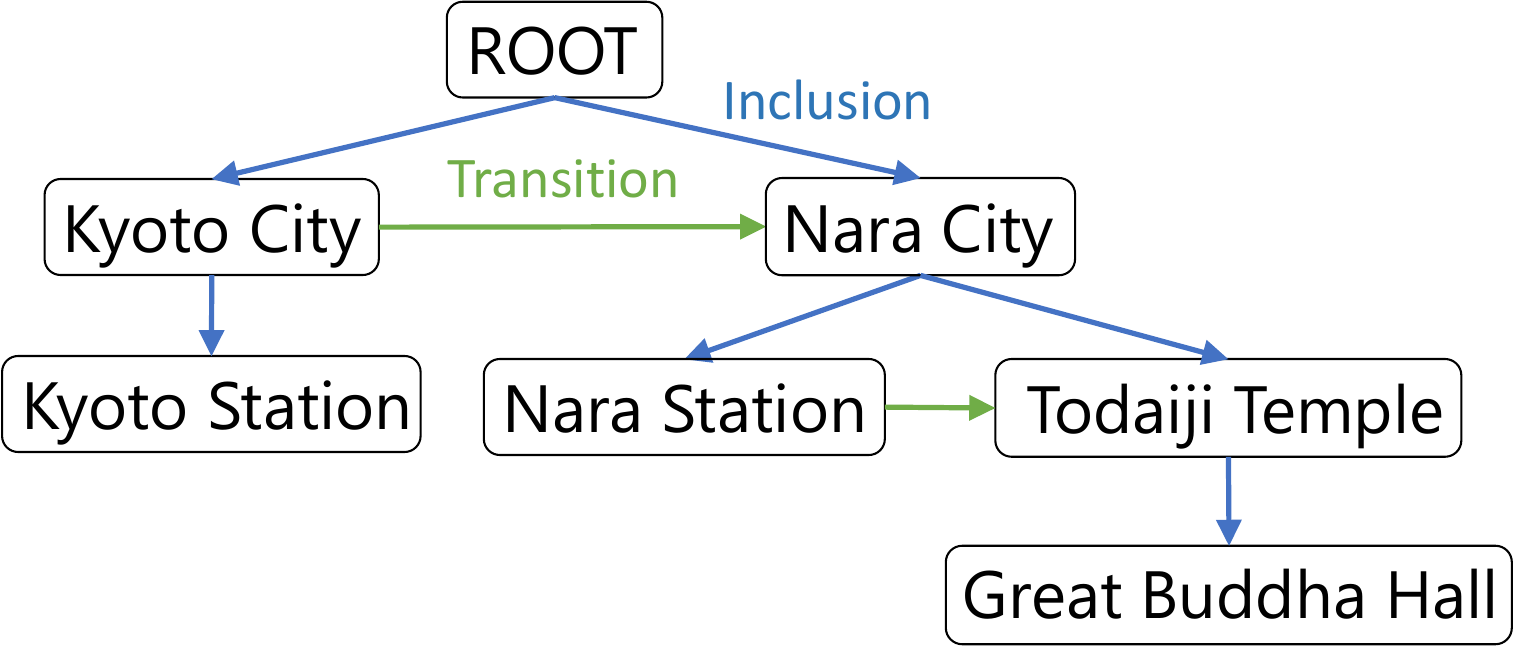}
  \caption{Example of a visiting order graph, the same example at the bottom of Figure~\ref{fig:step4_annotation_overview}.}
  \label{fig:annotation_order}
\end{figure}

We introduce a visiting order graph to represent 
non-linear relations of visited locations.
Figure~\ref{fig:annotation_order} shows an example.
Nodes correspond to entities, and edges correspond to relations between entities.
In Figure~\ref{fig:annotation_order}, blue vertical/diagonal directed edges are inclusion relations indicating the starting node geographically includes the ending node, and green horizontal directed edges are transition relations indicating the traveler visited the starting node entity and then directly visited the ending node entity, without visiting any other entities in between.

\paragraph{Inclusion Relation} \label{sec:def_inc}
Consider the example document in Figure~\ref{fig:step4_annotation_overview}, which describes that the traveler visited both ``Nara City'' and ``Todaiji Temple.''
Based on the geographic fact that the region of ``Nara City'' includes the region of ``Todaiji Temple,'' a reasonable interpretation is that the traveler visited the temple and thereby also visited the city simultaneously.
We introduce inclusion relation $\langle e_1, e_2 \rangle$, where an entity $e_1$ geographically includes another $e_2$.
We pick up the following two instances from the graph in Figure~\ref{fig:annotation_order}:
\begin{description}
    \setlength{\parskip}{0cm} 
    \setlength{\itemsep}{0cm}
    \item[] $p_1= \langle \textrm{Nara~City}, \textrm{Todaiji~Temple} \rangle$,
    \item[] $p_2= \langle \textrm{Todaiji~Temple}, \textrm{Great~Buddha~Hall} \rangle$.
\end{description}
Here, $p_1$ indicates that ``Nara City'' geographically includes ``Todaiji Temple'', and $p_2$ indicates that ``Todaiji Temple'' includes ``Great Buddha Hall.''
From these two relation instances, we can deduce a hierarchical relation that ``Nara City'' is a grandparent of ``Great~Buddha~Hall.''

\paragraph{Transition Relation} \label{sec:def_trans}

Given a set of entities for a document and inclusion relations among them, we assign transition relation to each pair of preceding and subsequent visited entities.
Notably, we restrict an entity pair with transition relation to two entities with the same parent entity.
In Figure~\ref{fig:annotation_order}, whereas ``Nara Station'' and ``Todaiji Temple'' have the same parent node, ``Kyoto Station'' and ``Nara Station'' do not.
Therefore, the transition relation can be assigned to $\langle \textrm{Nara Station}, \textrm{Todaiji Temple} \rangle$, but cannot be assigned to $\langle \textrm{Kyoto Station}, \textrm{Nara Station} \rangle$. 
This restriction allows us to determine the order of visits for any entity pairs by traversing transition and inclusion relations, even if entity pairs are not directly related to each other.
For example, ``Kyoto~Station'' does not have transition relation to ``Nara~City,'' but we can infer that ``Kyoto~Station'' was visited before ``Nara~City'' because the parent ``Kyoto~City'' has transition relation to ``Nara City.''

\paragraph{Other Criteria}
Visiting order graphs with the two relation types above can represent various trajectories, but not all.
We further introduce the following criteria for exceptional cases.
\begin{itemize}
\setlength{\parskip}{0cm} 
\setlength{\itemsep}{0.1cm}
\item Multiple Visits: There can be entities revisited after passing through other entities. Such an entity should be split into sub-entities that include the corresponding mentions for each visit, and sub-entities are regarded as nodes in a graph instead of the original entity.
\item \texttt{UnknownTime}: There can be entities to which the timing of the visit is ambiguous. Such an entity should be assigned the \texttt{UnknownTime} label and is excluded from nodes in a graph.
\item \texttt{Overlap}: There can be two entities geographically overlapping each other, but one does not include the other, e.g., ``Honshu'' and ``Tokyo Prefecture.''\footnote{Honshu is the largest island that constitutes the mainland Japan. Tokyo Prefecture is a geopolitical region that consists of a part of Honshu and 11 islands. } Such two entities should be assigned the \texttt{Overlap} relation, and either entity can be selected as a representative node to be assigned {inclusion} and {transition} relations between it and other entities.
\end{itemize}

\subsection{Annotation Data Construction} \label{sec:vop_ann}

After the visit status annotation step, only entities with the \texttt{Visit} label were retained as the nodes of a visiting order graph and used as input for a successive relation annotation step.
In this step, annotators assigned relations between the entities using an online whiteboard service, Miro,\footnote{\url{https://miro.com/}} which can be served as an annotation tool with GUI. 

\paragraph{Inter-Annotator Agreement}
We requested two annotators to independently annotate the same five documents as those used for visit status annotation.
We then measured the IAA using F1-score.
The obtained F1 scores suggest the moderate or high agreement: 0.94 for inclusion, 0.74 for transition, and 0.85 for both.

\paragraph{Data Statistics}
The 95 documents already annotated with visit status were allocated among five annotators, and one annotator annotated each document.
The total became 100 documents, including the aforementioned five documents, with 1,863 inclusion relation and 1,506 transition relation instances in total.
The detailed statistics are shown in Table~\ref{tab:order_stats} in Appendix~\ref{sec:app_stat}.

\subsection{Task Definition}
The task of VOP involves two subtasks: Inclusion Relation Prediction (IRP) and Transition Relation Prediction (TRP).

\paragraph{Inclusion Relation Prediction}
Given a set of entities $\mathcal{E}$ in a document, IRP requires a system to determine the parent entity for each entity $e_q \in \mathcal{E}$ from the set of candidate entities $\mathcal{P}^{(q)}_\mathrm{cand} = \mathcal{E} \setminus \{ e_q \} \cup \{ \texttt{ROOT} \} $.
If $e \in \mathcal{P}^{(q)}_\mathrm{cand}$ is predicted as the parent entity for $e_q$, it means that $e$ includes $e_q$.
The pseudo parent node \texttt{ROOT} should be predicted when the entity of interest has no parent entities.

\paragraph{Transition Relation Prediction}
Given a set of entities $\mathcal{E}$ in a document, TRP requires a system to determine the entity subsequently visited for each entity $e_q \in \mathcal{E}$ from the candidate set $\mathcal{S}^{(q)}_\mathrm{cand}$ with the same parent as that of $e_q$:
    $\mathcal{S}^{(q)}_\mathrm{cand} = \{e_k \in \mathcal{E}\,|\, e_k \neq e_q \,|\, \mathrm{Par}(e_k)=\mathrm{Par}(e_q) \} \cup \{ \texttt{EOS} \}$.
Here, $\mathrm{Par}(e)$ represents the parent entity of $e$, and the pseudo subsequent node \texttt{EOS} (End of Sequence) represents that the entity of interest has no subsequent entities.

\subsection{Baseline System} \label{sec:vop_baseline}
The baseline systems adopt similar methods for the two subtasks.
Specifically, for IRP and TRP, we select the most probable entity as the parent entity $\hat{e_p}$ and the subsequent entity $\hat{e_s}$ from the corresponding candidate set based on score functions $\mathrm{score}_\mathrm{par}$ and $\mathrm{score}_\mathrm{sub}$, respectively:
\begin{eqnarray} 
    \hat{e_p} &=& \argmax_{e^\prime \in \mathcal{P}^{(q)}_\mathrm{cand}} \:\: \mathrm{score}_\mathrm{par}(e_q, e^\prime), \\\label{eq:vop_inc_inference}
    \hat{e_s} &=& \argmax_{e^\prime \in \mathcal{S}^{(q)}_\mathrm{cand}} \:\: \mathrm{score}_\mathrm{sub}(e_q, e^\prime). \label{eq:vop_ba_inference}
\end{eqnarray}

\paragraph{Model Implementation}
To implement models with the score functions for both IRP and TRP, we used \texttt{LukeForEntityPairClassification}~\cite{yamada-etal-2020-luke} in Hugging Face Transformers, which receives the input text and the positions (character offsets) of two mentions.
The input text is constructed by concatenating the two sentences containing representative mentions for the target entity $e_q$ and a candidate entity $e'$ and all the sentences that occur between them, in the order of their occurrence.\footnote{For example, when \ja{奈良市} `Nara City' and \ja{東大寺} `Todaiji Temple' in Japanese translation of the first three sentences in Figure~\ref{fig:step4_annotation_overview} are entities of interest, the input text is represented as ``\texttt{<s>}\ja{その日は、京都市を素通りして、\texttt{<ent>}奈良市\texttt{<ent>}に向かいました。}\texttt{</s>}\texttt{<s>}$\dots$\texttt{</s>}\texttt{<s>}\ja{奈良駅で降りた後、駅から\texttt{<ent2>}東大寺\texttt{<ent2>}まで少し歩きました。}\texttt{</s>}''.}

\paragraph{Representative Mention Selection}
For IRP, proper noun mentions are prioritized over others and selected as representative ones.
For TRP, we define the priority for visit status labels, i.e., \texttt{Visit}\,$>$\,\texttt{See}\,$>$\,other labels, and select the mentions with higher priority as representative ones.\footnote{For both subtasks, preceding mentions are prioritized if there are multiple mentions with the same priority.}

\paragraph{Sequence Sorting Decoding}
For TRP, all sibling nodes sharing the same parent node should be arranged in a single sequence.
However, na\"{i}ve decoding based on predicted pairwise scores (Equation~\ref{eq:vop_ba_inference}) does not necessarily satisfy this requirement.
To address this issue, we introduce sequence sorting decoding based on a greedy search strategy as follows.
\begin{enumerate}
    \setlength{\parskip}{0cm} 
    \setlength{\itemsep}{0.1cm}
    \item Let $\mathcal{P}$ a set of all possible pairs whose nodes have the same parent. 
    \item Select a pair $\langle e_a, e_b \rangle$ with the highest score from $\mathcal{P}$.
    \item Exclude all pairs conflicting with $\langle e_a, e_b \rangle$ from $\mathcal{P}$, namely, $\langle e_b, e_a \rangle$, $\langle *, e_b \rangle$ and $\langle e_a, * \rangle$.
    Here, ``$*$'' indicates an arbitrary node.
    \item If transition relations among all the nodes have been determined, terminate the decoding. Otherwise, return to Step~2.
\end{enumerate}

\section{Experiments}
We evaluated the performance of our baseline systems and basic rule-based systems for the visit status prediction (VSP) task (\S\ref{sec:vsp_baseline}) and the visiting order prediction subtasks: inclusion relation prediction (IRP) and transition relation prediction (TRP) (\S\ref{sec:vop_baseline}).

\subsection{Experimental Settings} \label{sec:exp_setup}
\paragraph{Data Split}
As shown in Table~\ref{tab:stats}, we split the 100 documents in ATD-VSO into training, development, and test sets at a ratio of 7:1:2.

\paragraph{Task Settings}
We adopted the settings where gold labels of preceding tasks were given to systems, and evaluated the systems for each task independently. 
That is, systems take gold entities for VSP, the gold entities with gold visit status labels for IRP, and gold visited (sub-)entities\footnote{For TRP, we adopt a setting where the number of visits for each entity is given.} and gold inclusion relations for TRP, respectively, as inputs.
The main reason for the gold label settings for all the tasks is that we focus on investigating the system performance for each task independently.
We leave more advanced investigations, including evaluation of end-to-end models, for future work.

\paragraph{Evaluation Metrics}
For VSP, we measured the accuracy of predicted labels for input entities as well as macro-averaged F1 score over all labels.
For IRP, we measured the F1 score for extracting inclusion entity pairs from input entities.
For TRP, we measured the F1 score for extracting transition entity pairs, excluding pairs where the subsequent entity is \texttt{EOS}.

\paragraph{Model Training}
We constructed our baseline system by fine-tuning a pretrained model with the training set for each task.
Specifically, we used a pretrained multilingual LUKE~\cite{ri-etal-2022-mluke} model\footnote{\url{https://huggingface.co/studio-ousia/mluke-large-lite}} for VSP and the same pretrained Japanese LUKE~\cite{yamada-etal-2020-luke} model\footnote{\url{https://huggingface.co/studio-ousia/luke-japanese-base}} for IRP and TRP.
Unless otherwise specified, we report the mean accuracy or F1 score on the test set of five runs with different random seed values for the baseline system for each task.
More details on model training settings are shown in Appendix~\ref{sec:app_para}.

\subsection{Results for Visit Status Prediction}
\label{sec:res_vsp}

\begin{table}[t]
\centering
\small
\begin{tabular}{llcc}
\toprule
Level & Method & Acc. & Macro F1\\
\midrule
\multirow{2}{*}{Mention}
& Majority Label & 0.679 & 0.135 \\ 
& LUKE & \textbf{0.789} & \textbf{0.468}\\ 
\midrule
\multirow{2}{*}{Entity}
& Majority Label & 0.823 & 0.451 \\ 
& LUKE + MLA  & \textbf{0.862} & \textbf{0.740} \\ 
\bottomrule
\end{tabular}
\caption{System performance for visit status prediction.}
\label{tab:result_gold_span}
\end{table}

\paragraph{Systems}
We evaluated a rule-based system (Majority Label) and baseline systems (LUKE and LUKE+MLA).
The Majority Label rule always outputs the most frequent label, \texttt{Visit}, for both mention and entity levels.
LUKE indicates the baseline system, and MLA indicates the mention label aggregation rule described in~\S\ref{sec:vsp_baseline}.

\paragraph{Results} 
Table~\ref{tab:result_gold_span} shows the performance for mention-level and entity-level VSP.
Majority Label achieved good accuracy (0.679 for mentions and 0.823 for entities), indicating the imbalance in label distribution with a majority of \texttt{Visit} instances.
This aligns with the intuition that most of the locations mentioned in travelogues are visited ones.
The baselines, i.e., LUKE and LUKE+MLA, achieved better performance, owing to the capability of modeling both mention and context information, as discussed in detail in Appendix~\ref{sec:app_exp_vsp} (Table~\ref{tab:vop_result_mask}).

\begin{table}[t]
\centering
\small
\begin{tabular}{lllllll}
\toprule
\multirow{2}{*}{Label} &  \multicolumn{3}{c}{Mention} & \multicolumn{3}{c}{Entity} \\
& P & R & F1 & P & R & F1 \\
\midrule
\texttt{Visit}  & .835 & .913 & .872 & .896 & .942 & .918\\ 
\texttt{Plan}   & .706 & .688 & .696 & -- & -- & --\\
\texttt{See}    & .655 & .661 & .657 & -- & -- & --\\
\texttt{Past}   & 0    & 0    & 0    & -- & -- & --\\
\texttt{Future} & 0    & 0    & 0    & -- & -- & --\\
\texttt{UN/O}   & .611 & .403 & .482 & .650 & .495 & .561 \\
\bottomrule
\end{tabular}
\caption{Precision (P), Recall (R), and F1 scores of LUKE (mention-level) and LUKE+MLA (entity-level) for each label of visit status prediction.}
\label{tab:vsp_result_each_label}
\end{table}

\paragraph{Label-Wise Performance} Table~\ref{tab:vsp_result_each_label} shows the performance of the LUKE baselines for each label.
First, the baselines achieved high performance (F1 of 0.872--0.918) for \texttt{Visit} for both levels.
Second, the baselines resulted in limited performance (F1 of 0.482--0.561) for \texttt{UnkOrNotVisit}/\texttt{Other}.
The confusion matrices (Figure~\ref{fig:vso_cm_mention} and~\ref{fig:vso_cm_entity} in Appendix~\ref{sec:app_exp_vsp}) shows 
the most frequent errors were misclassification of \texttt{UnkOrNotVisit}/\texttt{Other} as \texttt{Visit}.

\paragraph{Directions for Improvement}
The LUKE baselines take limited context as input.
LUKE predicts a label for each mention based on the context within the sentence where the mention occurs, and the MLA rule predicts a label for the entity by merely aggregating the mention-level predictions.
A possible direction to improve entity-level performance is to construct an end-to-end model that integrates information on all mentions for an entity with the wider context of the document.


\begin{table}[t]
\centering
\small
\begin{tabular}{rrccc}
\toprule
\multirow{2}{*}{Depth} & \multirow{2}{*}{\#Ent} & \multicolumn{3}{c}{F1} \\
& & Random & Flat & LUKE \\
\midrule
All & 468 & 0.043 & 0.244 & \textbf{0.355} \\
\midrule
1 & 114 & 0.057 & \textbf{1} & 0.058 \\
$\geq$2 & 354 & 0.038 & 0 & \textbf{0.425} \\
\bottomrule
\end{tabular}
\caption{System performance (F1 score) for inclusion relation prediction. Depth indicates the distance to \texttt{ROOT} of entity nodes based on the gold inclusion hierarchy.}
\label{tab:result_inc_pred}
\end{table}


\subsection{Results for Inclusion Relation Prediction} \label{sec:results_irp}

\paragraph{Systems}
We evaluated two rule-based systems (Random and Flat) and a baseline system (LUKE).
Random is a rule-based system that randomly selects the parent entity from the candidate set for each entity. 
Flat is the other rule-based system that always selects \texttt{ROOT} as the parent entity for an arbitrary entity.
LUKE indicates the baseline system described in \S\ref{sec:vop_baseline}.

\paragraph{Results}
Table~\ref{tab:result_inc_pred} shows the performance (F1 score) of the evaluated systems for IRP.
Flat exhibited the better performance than Random (F1 of 0.244 vs 0.043), suggesting that predicting \texttt{ROOT} can be a reasonable strategy when systems do not have knowledge for specific entities.
LUKE achieved the best performance (F1 of 0.355) and, in particular, outperformed Random by a large margin for the entities whose gold parents are entities other than  \texttt{ROOT} (depth$\geq$2).
By contrast, LUKE showed poor performance similar to Random for the entities whose gold parents are \texttt{ROOT} (depth$=$1).
Detailed performance for each depth is shown in Table~\ref{tab:irp_result_detail} in Appendix~\ref{sec:app_irp}.

\paragraph{Directions for Improvement}
The current LUKE baseline has two limitations.
First, the absolute overall performance (F1 of 0.355) has substantial room for improvement.
Potential reasons for the limited performance are that (1) the pretrained LUKE model for general entity analysis tasks did not learn geographic relations among specific geo-entities, and (2) it was difficult to obtain generalized knowledge on geographic relations between entities from fine-tuning only with text-based features. 
Possible solutions include pretraining with geospatial information like GeoLM~\cite{li-etal-2023-geolm}, and fine-tuning a model with geocoding-based features, such as predicted coordinates and shapes of entities.
Second, the performance is quite low for entities whose parent is \texttt{ROOT}.
This is because the baseline predicts \texttt{ROOT} as the parent for an entity only when it predicts all candidate entities as non-parent. This may be improved by a method that can directly predict \texttt{ROOT} by assigning a vector representation to \texttt{ROOT} based on a dummy sentence.

\subsection{Results for Transition Relation Prediction}
\label{sec:results_trp}
\paragraph{Systems}
We evaluated three rule-based systems (Random and two variants of OccOrder) and two variants of the LUKE baseline system.
The Random rule randomly lines up candidate entities for each set of entities with the same parent entity.
OccOrder arranges candidate entities in the order of occurrence of each representative mention in their document; whereas the ``early mention'' strategy (OccOrder-EM) uses the earliest occurrence mention as the representative mention, the ``visit status'' strategy (OccOrder-VS) prioritizes mentions based on visit status label similarly to LUKE (\S\ref{sec:vop_baseline}).
The LUKE variants correspond to the baseline system with na\"{i}ve score-based decoding and sequence sorting decoding (\S\ref{sec:vop_baseline}).

\begin{table}[t]
\centering
\small
\begin{tabular}{llll}
\toprule
Method & All & Fwd. & Rev. \\
\midrule
Random & 0.190 & 0.247 & 0.061 \\
OccOrder (early mention) & 0.730 & 0.773 & 0 \\
OccOrder (visit status) & \textbf{0.758} & 0.794 & \textbf{0.386} \\
LUKE (na\"{i}ve score) & 0.680 & 0.737 & 0.298 \\
LUKE (sequence sorting) & 0.748 & \textbf{0.796} & 0.366\\ 
\bottomrule
\end{tabular}
\caption{System performance (F1 score) for transition relation prediction. ``All'' indicates the performance for all entities. ``Fwd.'' and ``Rev.'' indicate the performance for entities whose gold subsequent entities occurred after or before the entities of interest in documents, respectively, regarding their earliest mentions.}
\label{tab:result_tra_pred}
\end{table}

\begin{table*}[t]
\centering
\footnotesize
\begin{tabular}{l|p{6.75cm}|p{6.75cm}}
\toprule
SentID & Original Japanese Text & English Translation \\
\midrule
025-02 & \ja{今回は、\ulblue{松江}\,{\scriptsize ${}^{\textsc{G:}\texttt{UN}}_{\textsc{P:}\texttt{Vis}}$}\,と\ulblue{米子}\,{\scriptsize ${}^{\textsc{G:}\texttt{UN}}_{\textsc{P:}\texttt{Vis}}$}\,は素通りします。} & This time, I will skip \ulblue{Matsue}\,{\scriptsize ${}^{\textsc{G:}\texttt{UN}}_{\textsc{P:}\texttt{Vis}}$}\, and \ulblue{Yonago}\,{\scriptsize ${}^{\textsc{G:}\texttt{UN}}_{\textsc{P:}\texttt{Vis}}$}\,.\\
\vspace{-0.5pc} & \\
027-01 & \ja{\ulblue{松江しんじ湖温泉駅}\,{\scriptsize ${}^{\textsc{G:}\texttt{UN}}_{\textsc{P:}\texttt{Vis}}$}\,は終着駅です。} & \ulblue{Matsue Shinjiko Onsen Station}\,{\scriptsize ${}^{\textsc{G:}\texttt{UN}}_{\textsc{P:}\texttt{Vis}}$} is the final station. \\
\midrule
046 & \ja{この日は\ulblue{名古屋}\,{\scriptsize ${}^{\textsc{G:}\texttt{Plan}}_{\textsc{P:}\texttt{Plan}}$}\,で1泊する予定だったので、まだ早いけれど\ulblue{伊勢}\,{\scriptsize ${}^{\textsc{G:}\texttt{Vis}}_{\textsc{P:}\texttt{Vis}}$}\,を後にしました。} & I planned to stay one night in \ulblue{Nagoya}\,{\scriptsize ${}^{\textsc{G:}\texttt{Plan}}_{\textsc{P:}\texttt{Plan}}$}\,, so I left \ulblue{Ise}\,{\scriptsize ${}^{\textsc{G:}\texttt{Vis}}_{\textsc{P:}\texttt{Vis}}$}\, even though it was still early. \\
\midrule
024 & \ja{霊廟のある\ulblue{大雄寺}\,{\scriptsize ${}^{\textsc{G:}\texttt{Vis}}_{\textsc{P:}\texttt{Vis}}$}\,。} & Here is \ulblue{Daiouji Temple}\,{\scriptsize ${}^{\textsc{G:}\texttt{Vis}}_{\textsc{P:}\texttt{Vis}}$}\, with its mausoleum. \\
025 & \ja{\ulblue{駅}\,{\scriptsize ${}^{\textsc{G:}\texttt{Vis}}_{\textsc{P:}\texttt{Vis}}$}\,から距離があるのでタクシーに乗りました。} & I took a taxi because it was far from \ulblue{the station}\,{\scriptsize ${}^{\textsc{G:}\texttt{Vis}}_{\textsc{P:}\texttt{Vis}}$}\,. \\
\bottomrule
\end{tabular}
\caption{Actual sentences in documents: ID 23251 (top), 20816 (middle), and 20851 (bottom). The superscript~``G'' to \ulblue{mentions} indicates gold visit status labels and the subscript~``P'' indicates LUKE's prediction (\texttt{Vis} for \texttt{Visit}, \texttt{Plan} for \texttt{PlanToVisit}, and \texttt{UN} for \texttt{UnkOrNotVisit}).}
\label{tab:analysis_examples}
\end{table*}

\paragraph{Results}
Table \ref{tab:result_tra_pred} shows the performance of the evaluated systems for TRP.
For all pairs, the OccOrder variants achieved relatively high performance (F1 of 0.730--758). 
This matches the intuition that the order of locations being described in text corresponds with the order of locations being visited to some extent.
LUKE with the decoding based on its raw pairwise scores underperformed the OccOrder variants.
However, LUKE with the sequence sorting decoding improved approximately 0.07 F1 points and achieved performance on par with that of OccOrder-VS.
This result shows the effectiveness of the sequence sorting decoding.
We investigated another aspect related to the prediction difficulty: the size of the candidate entity set might have an impact on the drop of the Random performance as observed in Table~\ref{tab:trp_result_detail} in Appendix~\ref{sec:app_trp}. 
However, no clear tendencies were observed for OccOrder and LUKE.

\paragraph{Directions for Improvement}
The current LUKE baseline has two limitations.
First, the vector representation of an entity is constructed from a single mention selected by the heuristic rule (\S\ref{sec:vop_baseline}), which limits the context of the entity. 
This would be improved by extending the context to include all mentions for two entities of interest, although it is necessary to explore an effective method that can grasp complicated relations among many mentions.
Second, the current baseline uniformly treats entity pairs without transition relation as negative instances.
However, entity pairs with indirect transition relation, where one is visited before the other via one or more entities, can be exploited as positive instances for an auxiliary task, similarly to relative event time prediction~\cite{wen-ji-2021-utilizing}.


\begin{figure*}
\centering
\includegraphics[width=16cm]{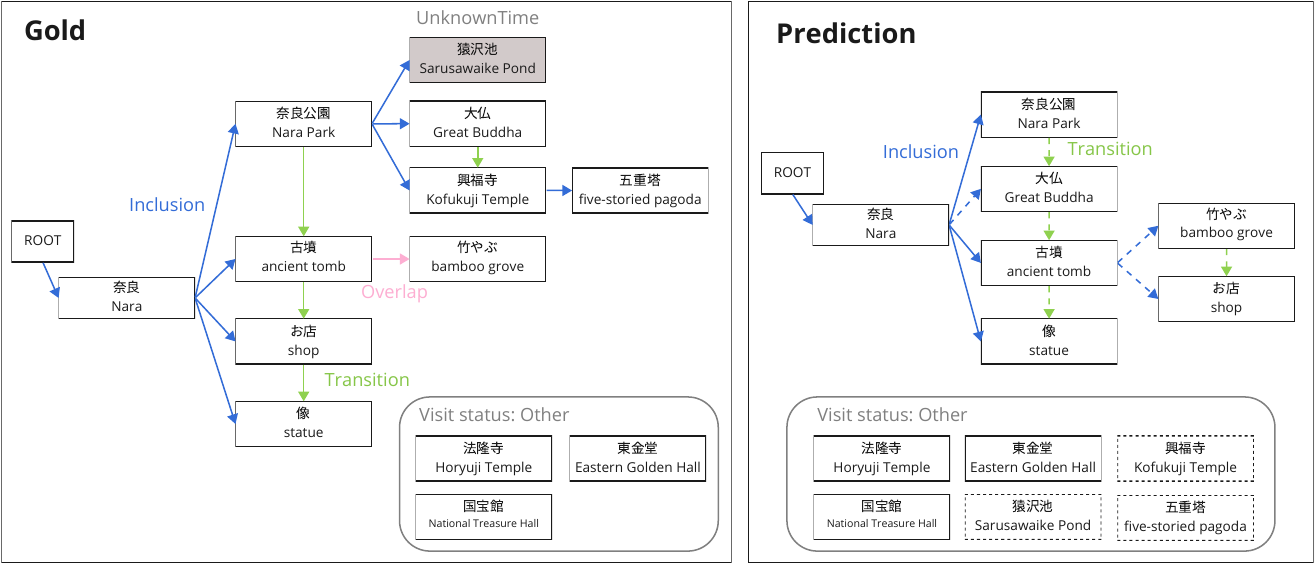}
\caption{Illustrations of the gold visiting order graph and LUKE's prediction for an actual document (ID 00019). The nodes with dashed frames and the edges with dashed arrows represent prediction errors.}
\label{fig:pipeline_example}
\end{figure*}
\begin{table*}[th]
\centering
\footnotesize
\begin{tabular}{l|p{6.75cm}|p{6.75cm}}
\toprule
SentID & Original Japanese Text & English Translation \\
\midrule
005 & \ja{\ulblue{大仏}\,{\scriptsize ${}^{\textsc{G:}\texttt{Vis}}_{\textsc{P:}\texttt{Vis}}$}\,様はとっても大きかったなぁ\textasciitilde} & \ulblue{The Great Buddha}\,{\scriptsize ${}^{\textsc{G:}\texttt{Vis}}_{\textsc{P:}\texttt{Vis}}$}\, was really huge.\\
\vspace{-0.5pc} & \\
009 & \ja{写真は\ulblue{猿沢池}\,{\scriptsize ${}^{\textsc{G:}\texttt{UN}}_{\textsc{P:}\texttt{See}}$}\,からも見える \ulblue{興福寺}\,{\scriptsize ${}^{\textsc{G:}\texttt{Vis}}_{\textsc{P:}\texttt{See}}$}\,の\ulblue{五重塔}\,{\scriptsize ${}^{\textsc{G:}\texttt{Vis}}_{\textsc{P:}\texttt{See}}$}\,です。} 
& It's a photo of \ulblue{the five-storied pagoda}\,{\scriptsize ${}^{\textsc{G:}\texttt{Vis}}_{\textsc{P:}\texttt{See}}$}\, at \ulblue{Kofukuji Temple}\,{\scriptsize ${}^{\textsc{G:}\texttt{Vis}}_{\textsc{P:}\texttt{See}}$}\, visible from \ulblue{Sarusawaike Pond}\,{\scriptsize ${}^{\textsc{G:}\texttt{Vis}}_{\textsc{P:}\texttt{See}}$}\,.\\
\vspace{-0.5pc} & \\
018 & \ja{\ulblue{古墳}\,{\scriptsize ${}^{\textsc{G:}\texttt{Vis}}_{\textsc{P:}\texttt{Vis}}$}\,の中に入ると、さらに大きさを感じることができます。} & When entering \ulblue{the ancient tomb}\,{\scriptsize ${}^{\textsc{G:}\texttt{Vis}}_{\textsc{P:}\texttt{Vis}}$}\,, you can perceive the size more clearly. \\
\vspace{-0.5pc} & \\
019 & \ja{\ulblue{竹やぶ}\,{\scriptsize ${}^{\textsc{G:}\texttt{Vis}}_{\textsc{P:}\texttt{Vis}}$}\,の中にひっそりとあります。} & It is quietly situated in \ulblue{a bamboo grove}\,{\scriptsize ${}^{\textsc{G:}\texttt{Vis}}_{\textsc{P:}\texttt{Vis}}$}\,.\\
\vspace{-0.5pc} & \\
021 & \ja{「柿の葉寿司」で有名な\ulblue{お店}\,{\scriptsize ${}^{\textsc{G:}\texttt{Vis}}_{\textsc{P:}\texttt{Vis}}$}\,です。} & This is \ulblue{a shop}\,{\scriptsize ${}^{\textsc{G:}\texttt{Vis}}_{\textsc{P:}\texttt{Vis}}$}\, famous for its persimmon-leaf sushi.\\
\bottomrule
\end{tabular}
\caption{
Actual sentences in document 00019 with gold visit status (``G'') and that prediced by LUKE (``P'').}
\label{tab:analysis_examples2}
\end{table*}

\section{Error Analysis}
\label{sec:error_analysis}

We investigated tendencies of prediction errors of the current baseline systems: LUKE+MLA for VSP, LUKE for IRP, and LUKE with na\"{i}ve score decoding (\S\ref{sec:analysis_trp}) or sequence sorting decoding (\S\ref{sec:analysis_pipe}) for TRP.
The following subsections describe our error analysis in the settings of independent prediction for each task and integrated pipeline prediction for the three tasks.

\subsection{Visit Status Prediction}
As stated in \S\ref{sec:res_vsp}, the LUKE baselines often misclassified mentions with the \texttt{UnkOrNotVisit} label and entities with the \texttt{Other} label.
Our analysis reveals two error tendencies.
First, LUKE sometimes failed to distinguish factual statements from descriptions of traveler's visitation.
For example, the sentence 027-01 in Table~\ref{tab:analysis_examples} (document 23251) is a factual statement, not indicating the visitation, so the mention ``Matsue Shinjiko Onsen Station'' should have been assigned \texttt{UnkOrNotVisit}.
Second, LUKE sometimes fails to correctly understand the meaning of some motion verbs, such as ``skip'' and ``pass on.''
For example, the mentions ``Matsue'' and ``Yonago'' are clearly stated as locations that the writer did not visit, as shown in Table~\ref{tab:analysis_examples} (document 23251, sentence 025-02).

\subsection{Inclusion Relation Prediction}
\label{sec:analysis_irp}
The results shown in Table~\ref{tab:result_inc_pred} (\S\ref{sec:results_irp}) have indicated that IRP is a challenging task.
Our analysis reveals that LUKE learned the tendency that prefectures and cities often become parents of some entities, but LUKE also sometimes made incorrect predictions, such as a prefecture/city being the parent of another prefecture/city.
For example, LUKE predicted ``Nagoya'' as the parent of ``Ise,'' although both are cities (sentence 046 of document 20816 in Table~\ref{tab:analysis_examples}).
This suggests that the model lacks commonsensical geographic knowledge.



\subsection{Transition Relation Prediction}
\label{sec:analysis_trp}
The results shown in Table~\ref{tab:result_tra_pred} (\S\ref{sec:results_trp}) have indicated difficulty in predicting reverse-order entity pairs.
An example reverse pair is in sentences 024 and 025 (document 20851) in Table~\ref{tab:analysis_examples}.
While ``Daiouji Temple'' precedes ``the station,'' these sentences describe that the traveler moved from the station to the temple.
Although LUKE tended to predict the correct order of reverse pairs when there were some clues, such as temporal expressions like ``before'' and ``after,'' but the system made incorrect predictions for reverse pairs without salient clues, including the above example.


\subsection{Pipeline Prediction}
\label{sec:analysis_pipe}

Figure~\ref{fig:pipeline_example} shows the gold visiting order graph and prediction from the baseline systems\footnote{They correspond to the systems stated at the beginning of \S\ref{sec:error_analysis}. We simply refer to each system as ``LUKE'' here.} connected in the pipeline for document 00019.

For VSP, LUKE made correct predictions for 10 out of 13 entities.
The three errors were caused by the predictions for three mentions in sentence 009 in Table~\ref{tab:analysis_examples2}; LUKE incorrectly predicted the mention label \texttt{See}, and then the MLA rule determined the entity label \texttt{Other}.
LUKE failed to grasp the nuanced meaning of the sentence, which describes a photo of the facilities (``five-storied pagoda'' and ``Kofukuji Temple'') taken by the traveler and the nearby location (``Sarusawaike Pond'').

For IRP, LUKE predicted correct parents for four out of seven entities with the predicted label \texttt{Visit} and incorrect parents for the remaining three entities.
Two of the failed entities are written with general noun mentions (``bamboo grove'' in sentence 019 and ``shop'' in sentence 021), and the geographic relations regarding these locations are not explicitly described.
Therefore, the parents of them (``Nara'') should be determined based on the consistent context of the traveler's trip to Nara.
For correct prediction for another failed entity regarding the mention ``Great Buddha'' in sentence 005, which refers to famous Birushana Buddha at Todaiji Temple, geographic knowledge that Todaiji Temple is located in Nara Park is also necessary.

For TRP, LUKE was able to identify no exact entity pairs with correct transition relation mainly because the incorrect inclusion hierarchy was given as an input.
This result indicates that the more accurate prediction of inclusion relation is crucial for the accurate prediction of transition sequences.
However, LUKE successfully predicted many indirect transition relations, for example, ``Nara Park,'' ``ancient tomb,'' and ``stature'' have been visited in this order as LUKE also predicted.
This suggests more lenient evaluation metric would also be useful in distinguishing such reasonable predictions from severe errors, such as pairs arranged in the reverse order from the correct visiting order.

\section{Related Work}
\label{sec:related_work}


\subsection{Visit Status Prediction}
Visit status prediction can be regarded as a task in the field of human movement analysis for text, which includes a major stream of \textit{predicate-centric} analysis and another \textit{location-centric} analysis.

\paragraph{Predicate-Centric Analysis}
Represented by \textsc{SpaceBank}, a line of studies have sought to develop computational models that can recognize, generate and reason about spatial information, including place names, topological relations, and human movement~\cite{pustejovsky-etal-2012-linguistically,pustejovsky-yocum-2013-capturing,pustejovsky-etal-2015-semeval}.
These studies basically focused on verbs as expressions of movement.
Similarly, previous event analysis studies from temporal or factuality perspectives treated verbs or predicates as a trigger of each event and specified attribute information on verbs, as represented by \textsc{TimeML}~\cite{pustejovsky2003timeml} and \textsc{FactBank}~\cite{sauri2009factbank}.
In user-generated contents including travelogues, geographic movements are often expressed without predicates, for example, \textit{scene transition} by changing paragraphs or posts.
Thus, we focus on geo-entities and their mentions, in contrast to the above predicate-centric analysis studies.

\paragraph{Location-Centric Analysis}
Some previous studies focused on ``mention-level'' visit status of location mentions in SNS posts~\cite{li-etal-2014-fine-grained-poi-recognition,matsuda2018monitoring} and clinical documents~\cite{peterson2021automated}, which can be regarded as location-centric analysis.
Long documents like travelogues often contain multiple mentions referring to the same location (i.e., geo-entity), and such different mentions can have different visit status labels.
This necessitates a step of aggregating various visit status labels of mentions into the final visit status of the entity, which indicates whether the traveler eventually visited the location or not. 
Therefore, we additionally address ``entity-level'' prediction.

\subsection{Visiting Order Prediction}
Compared to many existing studies on the extraction of geo-entity mentions or toponyms~\cite{lieberman-etal-2010-geotagging,matsuda-etal-2017-geographical,kamalloo-etal-2018,wallgrun-etal-2018-geocorpora,weissenbacher-etal-2019-semeval-geoparsing,gritta-etal-2020-pragmatic,higashiyama-etal-2023-atd}, a few studies extracted some types of geographic \textit{trajectories} from text~\cite{ishino-2012-extracting,wagner-etal-2023-event,kori-2006-blog}.

\citet{ishino-2012-extracting} proposed a task and a method to extract transportation information, i.e., the origin and destination with a transportation method, from each disaster-related tweet.
Such information can be regarded as part of a movement trajectory.
\citet{wagner-etal-2023-event} proposed a task to extract an entire trajectory from each transcription of testimony videos.
Each testimony segment, which typically corresponds to one minute of the original video, was assigned a predefined location category, such as ``\texttt{cities in Austria}'' and ``\texttt{ghettos in Hungary}.''
That is, their trajectory is a transition sequence of a coarse-grained location mainly mentioned in each scene, not a detailed movement trajectory of specific locations.
\citet{kori-2006-blog} proposed a method to extract and summarize sequences of location mentions as visitors' representative trajectories in blogs.
They determined the visiting order of locations based on the order that mentions occurred in blogs.
In contrast to \citet{kori-2006-blog}, we aim to identify the faithful visiting order, which aligns with the written intentions.
The crucial difference between these three studies and ours is the trajectory representation; we adopt not \textit{sequences} but \textit{graphs} as appropriate representations that retain geographic hierarchy as well as the temporal order of locations.

\section{Conclusion}
This study has proposed a \textit{visiting order graph} to represent non-linear relations of visited locations and constructed an annotated travelogue dataset for graph-structured trajectory extraction.
The experiments using the dataset have indicated the accuracy of baseline systems with some directions for performance improvement.
The error analysis of actual examples has revealed notable tendencies of the prediction errors.
One future direction is to develop sophisticated systems by addressing the issues we have raised for improvement.
Another direction is to develop an integrated system for both trajectory extraction and grounding, which extracts an trajectory from each input document and associates its locations with points/areas of a map.

\clearpage
\bibliography{tacl2021}
\bibliographystyle{acl_natbib}

\clearpage
\appendix

\section{Annotation Examples}
\label{sec:app_label_example}

\begin{table}[h]
\centering
\small
\begin{tabular}{llp{5.25cm}}
\toprule
1\!\! & \texttt{Visit} & \ja{無事に\ulblue{赤岳山頂}に着きました!}\\
&& (Arrived safely at the \ulblue{summit of Mount Akadake}!)\\
2\!\! & \texttt{Plan} & \ja{\ulblue{穂高神社}に向かいます。}\\
&& (Heading to \ulblue{Hotaka Shrine}.)\\
3\!\! & \texttt{See} & \ja{\ulblue{硫黄岳}が近くに見えて来ました。} (\ulblue{Mount Iodake} is becoming visible nearby.) \\
4\!\! & \texttt{Past} & \ja{3\textasciitilde 4年前に\ulblue{浪速餃子スタジアム}で} \\
&& (About 3–4 years ago at the \ulblue{Naniwa Gyoza
Stadium}) \\
5\!\! & \texttt{Future} & \ja{今度は、\ulblue{松江}と\ulblue{米子}に来てみたいものです。} (Next time, I'd like to visit \ulblue{Matsue} and \ulblue{Yonago}.) \\
6\!\! & \texttt{UnkNot} & \ja{\ulblue{糸魚川駅}行きの車両は1両です。} \\
&& (The train bound for \ulblue{Itoigawa Station} consists of one car.)\\
\bottomrule
\end{tabular}
\caption{Example mentions annotated with visit status labels. \texttt{Plan}, \texttt{Past}, \texttt{Future}, and \texttt{UnkNot} indicate \texttt{PlanToVisit}, \texttt{Visit-Past}, \texttt{Visit-Future}, and \texttt{UnkOrNotVisit}, respectively.}
\label{tab:visit_labels_example}
\end{table}

\section{Detailed Dataset Statistics}
\label{sec:app_stat}

\begin{table}[h]
\centering
\small
\begin{tabular}{lrrrrrr}
\toprule
Set & \texttt{Visit} & \texttt{Plan} & \texttt{See} & \texttt{Past} & \texttt{Fut} & \texttt{UN/O}\\
\midrule
Train \!\!& 1,942 & -- & -- & -- & -- & 397\\
Dev   &   252 & -- & -- & -- & -- &  64\\
Test  &   575 & -- & -- & -- & -- & 124\\
\midrule
Train \!\!& 2,577 & 358 & 212 & 10 & 6 & 619\\
Dev   & 332 & 48 & 46 & 1 & 4 & 74 \\
Test  & 748 & 121 & 59 & 10 & 4 & 160\\
\bottomrule
\end{tabular}
\caption{Numbers of visit status labels for entities (top) and mentions (bottom). \texttt{Plan}, \texttt{Past}, and \texttt{Fut} indicate \texttt{PlanToVisit}, \texttt{Visit-Past}, and \texttt{Visit-Future}, respectively. \texttt{UN/O} indicates \texttt{\textbf{U}nkOr\textbf{N}otVisit} for mention level and \texttt{\textbf{O}ther} for entity level.}
\label{tab:men_stats}
\end{table}

\begin{table}[h!]
\centering
\small
\begin{tabular}{lrrrrr}
\toprule
Set & \texttt{Inc} & \texttt{Trans} & \texttt{Over} & \texttt{UnkTime} & MV \\
\midrule
Train & 1,302 & 1,041 & 38 & 35 & 95\\
Dev   &   186 &   143 &  8 &  8 & 16\\
Test  &   375 &   322 &  5 & 10 & 32\\
\bottomrule
\end{tabular}
\caption{Detailed statistics for visiting order annotation. \texttt{Inc} (\texttt{Inclusion}), \texttt{Trans} (\texttt{Transition}), and \texttt{Over} (\texttt{Overlap}) indicate the numbers of entity pairs with each relation type. \texttt{UnkTime} (\texttt{UnknownTime}) indicates the number of entities with the label. MV indicates the number of entities with multiple visits.}
\label{tab:order_stats}
\end{table}

\newpage
\section{Model Training Settings}
\label{sec:app_para}

\begin{table}[h]
\centering
\small
\begin{tabular}{lcc}
\toprule
Parameter & VSP & VOP\\
\midrule
Training epochs & 10 & 10\\ 
Batch size & 16 & 8\\ 
Gradient accumulation step & 4 & 1 \\
Learning rate & 5e-6 & 5e-6 \\ 
\bottomrule
\end{tabular}
\caption{Hyperparameters of LUKE models for mention-level visit status prediction (VSP) and visiting order prediction (VOP).}
\label{tab:hyper_parameters}
\end{table}

\section{Detailed Experimental Results}
\label{sec:app_exp}

\subsection{Visit Status Prediction}
\label{sec:app_exp_vsp}

\begin{figure}[h]
    \centering
    \includegraphics[width=7.5cm]{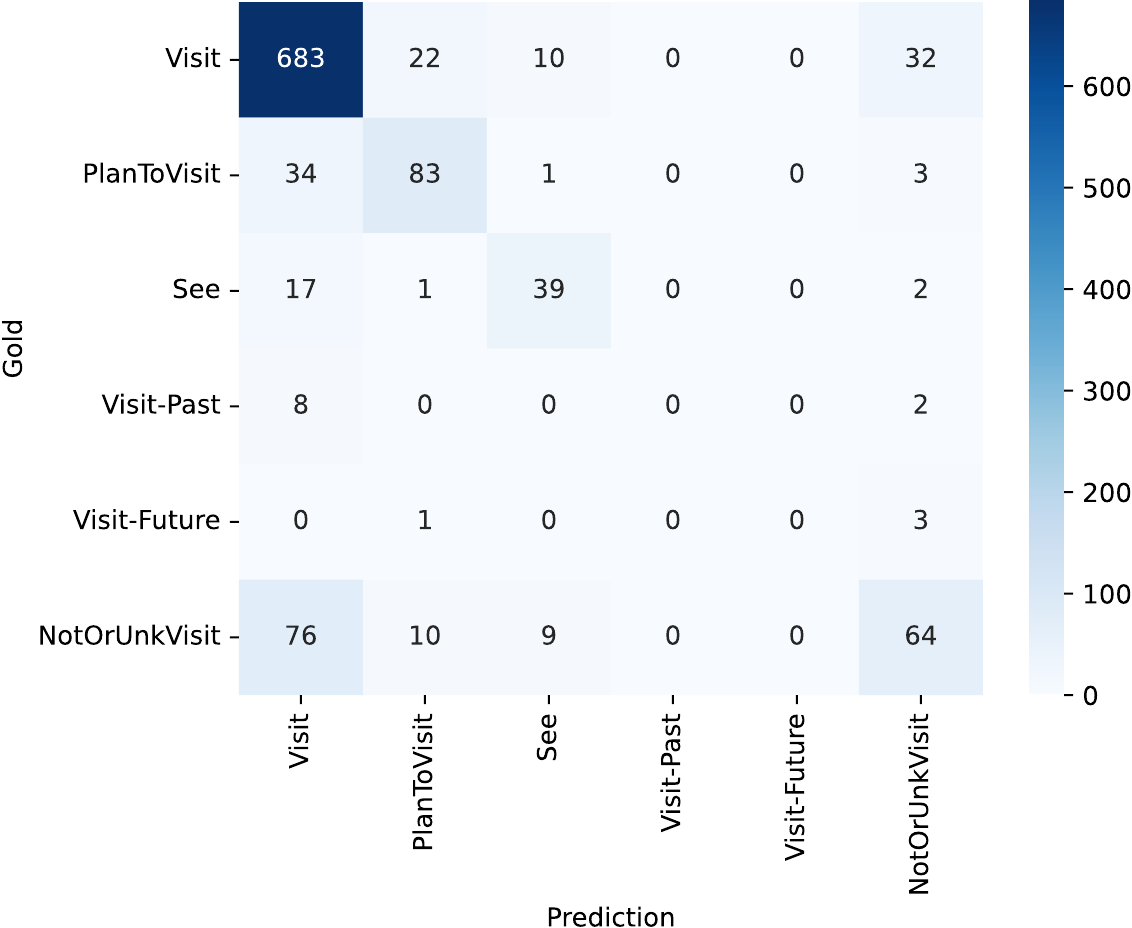}
    \caption{Confusion matrix of LUKE for mention-level visit status prediction.}
    \label{fig:vso_cm_mention}
\end{figure}
\begin{figure}[h]
    \centering
    \includegraphics[width=6.5cm]{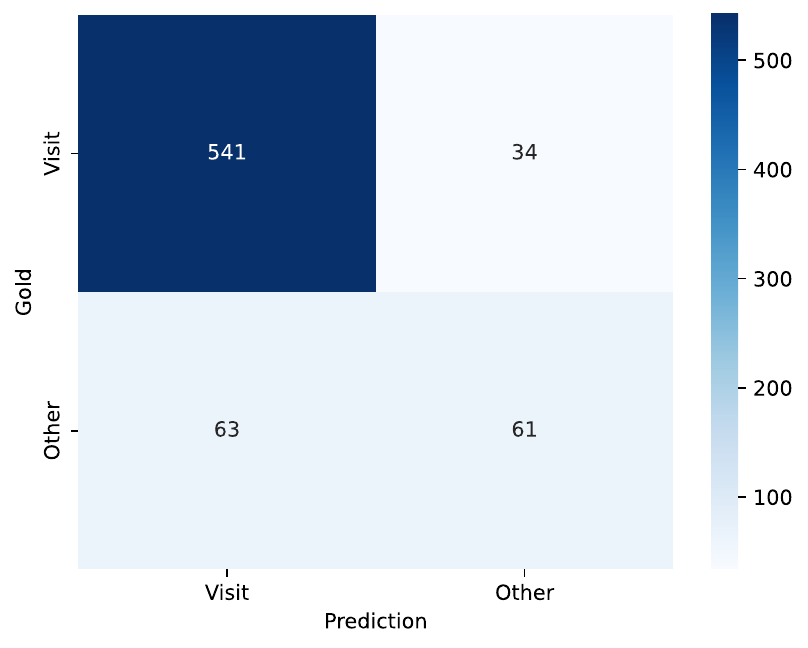}
    \caption{Confusion matrix of LUKE for entity-level visit status prediction.}
    \label{fig:vso_cm_entity}
\end{figure}



\paragraph{Ablation Study}
To investigate the influence of surface text on learning and prediction of the baseline model for mention-level VSP, we evaluated two additional variants of the LUKE baseline trained with edited input text.
That is, (1) ``mention masking'' model trained with input text where mention tokens are replaced by \texttt{[MASK]} tokens, and (2) ``mention only'' model trained with input text where context tokens are removed.
Table~\ref{tab:vop_result_mask} shows the performance of the model variants on the development set.
Compared to the original baseline, the mention masking model remained slightly lower in accuracy (-0.024 points), and the mention only model, while even lower in accuracy (-0.145 points), was still able to predict correct labels to some extent.
This suggests that the model mainly relied on context information and also used mention information together.

\begin{table}[h!]
\centering
\small
\begin{tabular}{lcc}
\toprule
Method & Acc. & Macro F1\\
\midrule
LUKE & 0.817 & 0.486\\ 
LUKE (mention masking) & 0.793 & 0.457\\ 
LUKE (mention only) & 0.672 & 0.205 \\ 
\bottomrule
\end{tabular}
\caption{Performance of LUKE variants for mention-level visit status prediction on the development set.}
\label{tab:vop_result_mask}
\end{table}

\subsection{Inclusion Relation Prediction}
\label{sec:app_irp}

\begin{table}[h]
\centering
\small
\begin{tabular}{rrccc}
\toprule
\multirow{2}{*}{Depth} & \multirow{2}{*}{\#Ent} & \multicolumn{3}{c}{F1} \\
& & Random & Flat & LUKE \\
\midrule
1 & 114 & 0.057 & \textbf{1} & 0.058 \\
2 & 194 & 0.040 & 0 & \textbf{0.432} \\
3 & 111 & 0.035 & 0 & \textbf{0.438} \\
4 & 42  & 0.034 & 0 & \textbf{0.305} \\
5 & 7   & 0.034 & 0 & \textbf{0.743} \\
\bottomrule
\end{tabular}
\caption{System performance (F1 score) for inclusion relation prediction for each depth. Depth indicates the distance to \texttt{ROOT} of entity nodes based on the gold inclusion hierarchy.}
\label{tab:irp_result_detail}
\end{table}



\newpage
\subsection{Transition Relation Prediction}
\label{sec:app_trp}

\begin{table}[h]
\centering
\small
\begin{tabular}{rccccc}
\toprule
\multirow{2}{*}{Size} & \multirow{2}{*}{\#} & \multicolumn{4}{c}{F1}\\
& & Rand. & \!Occ-V\! & \!LUKE-N\! & \!LUKE-S\! \\
\midrule
2 & 34 & 0.498 & \textbf{0.941} & 0.782 & 0.919 \\
3 & 36 & 0.332 & \textbf{0.861} & 0.744 & 0.845 \\
4 & 21 & 0.245 & 0.667 & 0.685 & \textbf{0.732} \\
5 & 24 & 0.196 & 0.750 & 0.754 & \textbf{0.808} \\
6 & 35 & 0.165 & \textbf{0.857} & 0.765 & 0.847 \\
7 & 12 & 0.138 & 0.750 & 0.435 & \textbf{0.517} \\
8 & 35 & 0.126 & \textbf{0.629} & 0.549 & 0.566 \\
9 & 32 & 0.108 & \textbf{0.812} & 0.718 & 0.800 \\
$\geq$10 & 93 & 0.069 & 0.667 & 0.627 & \textbf{0.681} \\
\bottomrule
\end{tabular}
\caption{System performance for transition relation prediction for each size of candidate entity sets. OccV represents OccOrder with the visit status strategy. LUKE-N and LUKE-S represent LUKE with na\"{i}ve score-based decoding and sequence sorting decoding, respectively.}
\label{tab:trp_result_detail}
\end{table}

\end{document}